\documentclass[letterpaper, 10 pt, conference]{ieeeconf}  

\IEEEoverridecommandlockouts                              

\usepackage{amsmath,amssymb}
\usepackage{graphicx}
\usepackage{cite}
\usepackage{url}

\title{\LARGE \bf
Hyperbolically-Discounted Reinforcement Learning on Reward-Punishment Framework
}

\author{Taisuke Kobayashi$^{1}$
\thanks{$^{1}$T. Kobayashi is with the Division of Information Science, Nara Institute of Science and Technology, 8916-5 Takayama-cho, Ikoma, Nara 630-0192, Japan
{\tt\small kobayashi@is.naist.jp}}%
}

\begin{document}

\maketitle
\thispagestyle{empty}
\pagestyle{empty}

\begin{abstract}

This paper proposes a new reinforcement learning with hyperbolic discounting.
Combining a new temporal difference error with the hyperbolic discounting in recursive manner and reward-punishment framework, a new scheme to learn the optimal policy is derived.
In simulations, it is found that the proposal outperforms the standard reinforcement learning, although the performance depends on the design of reward and punishment.
In addition, the averages of discount factors w.r.t. reward and punishment are different from each other, like a sign effect in animal behaviors.

\end{abstract}

\section{Introduction}

Reinforcement learning (RL) basically acquires the optimal policy so as to maximize a return, which accumulates future rewards with exponential discounting~\cite{suttonRLbook1998}.
Recent work has shown that RL is an excellent approach to achieve complicated tasks by robots~\cite{luoICRA2018,tsurumineRAS2019}.
The reason why the exponential discounting is used is that it is mathematically easy to handle with a recursive manner.

However, animals show behaviors, which cannot be explained when using the exponential discounting~\cite{kobayashiJNS2008}.
Specifically, immediate and small reward is preferred to future and large reward, but such a decision is reversed when a moderate offset (delay) is added to the time.
To explain such behaviors, a hyperbolic discounting is proposed in the context of behavioral economics.

It is natural to judge that the hyperbolic discounting has some advantages (e.g., the above decision making, long-tailed discounting of the future reward, etc.) if animals certainly use it.
Following this intuition, this paper aims to switch RL from with the exponential discounting to with the hyperbolic discounting.
To this end, a temporal difference (TD) error for learning is redefined with the hyperbolic discounting from the literature~\cite{alexanderNC2010}.
Its original definition, however, assumes that the value function (i.e., the expectatoin of the return) an reward are positive.
To handle real number of reward for generality, a reward-punishment framework in RL~\cite{okadaIWANN2001,elfwingICDL2017}, which divides real reward into positive one (called reward) and negative one (called punishment), is employed.
Note that this framework is also proposed from biological features.

The proposed method is simply investigated in numerical simulations.
The results imply that it can outperform the conventional RL depending on the design of reward and punishment.
In addition, it is found that the average discount factors for reward and punishment are in asymmetry, which is a similar feature to animals, called a sign effect~\cite{tanakaJNS2014}.

\section{Proposal}

\subsection{Hyperbolically-discounted temporal difference}

Let's define the value function (the return), $V$, with the hyperbolic discounting as follows:
\begin{align}
V_t = \mathbb{E}\left [ \sum_{k=0}^\infty \frac{r_{t+k}}{1 + \kappa k} \mid s_t \right ] \label{eq:value_hyp}
\end{align}
where $r$ and $s$ denote reward and state, respectively, and $\kappa$ is the hyperparameter related to the discount factor $\gamma$ ($\gamma = 1 -\kappa$ when the exponential discounting).

Actually, this equation is difficult to solve with the recursive manner.
The literature~\cite{alexanderNC2010}, however, has derived a hyperbolically-discounted TD error, $\delta$, as follows:
\begin{align}
\delta_t = r_t + \left (1 - \frac{\kappa V_t}{(\mu_r + b)^p} \right )V_{t+1} - V_t \label{eq:tde_hyp}
\end{align}
where the bias $b$ is given below in this paper.
\begin{align}
b = \beta \sigma_r \label{eq:bias_hdtd}
\end{align}
$\mu_r$ and $\sigma_r$ mean the statistics of reward (mean and standard deviation, respectively).
$\beta$ and $p$ are the hyperparameter.
Here, $\gamma$ is defined as $1 - \frac{\kappa V_t}{(\mu_r + \beta \sigma_r)^p}$.

Here, two problems in this definition are raised.
One is the difference of the scale between $r$ (and $\mu_r$) and $V$.
From the sum of geometric progression, $V$ would be $1/(1-\gamma)$ times larger than $r$ if $\gamma$ is constant.
To compensate this scale gap, reward to calculate TD error is multiplied with $1-\bar \gamma$ where $\bar \gamma$ the average discount factor.

\subsection{Reward-punishment framework}

Another problem is signs of numerator and denominator.
Reward in RL is defined as real number, so there is the possibility to get $\gamma > 1$ when $V$ and $\mu_r$ have the different signs.
To avoid this without loss of generality, the reward-punishment framework is employed.
Specifically, real reward is divided into positive one $r$, called reward, and negative one with inverted sign $p$, called punishment, in environment~\cite{okadaIWANN2001} side or agent side~\cite{elfwingICDL2017}.
In that case, both reward and punishment are defined to be positive real numbers, and therefore, $\gamma > 1$ would never be caused.

However, the value functions for reward and punishment, $V_r$ and $V_p$, are actually approximated using some function approximators (e.g., deep neural networks).
The approximated values should have correct domain, i.e., the positive real numbers, as follows:
\begin{align}
V_{r,p} = \max(0, y_{r,p})
\end{align}
where $y_{r,p}$ are the outputs of the function approximators.

\begin{figure*}[tb]
	\centering
	\begin{minipage}[t]{0.495\linewidth}
		\centering
		\begin{minipage}[t]{0.4925\linewidth}
			\centering
			\includegraphics[keepaspectratio=true,width=\linewidth]{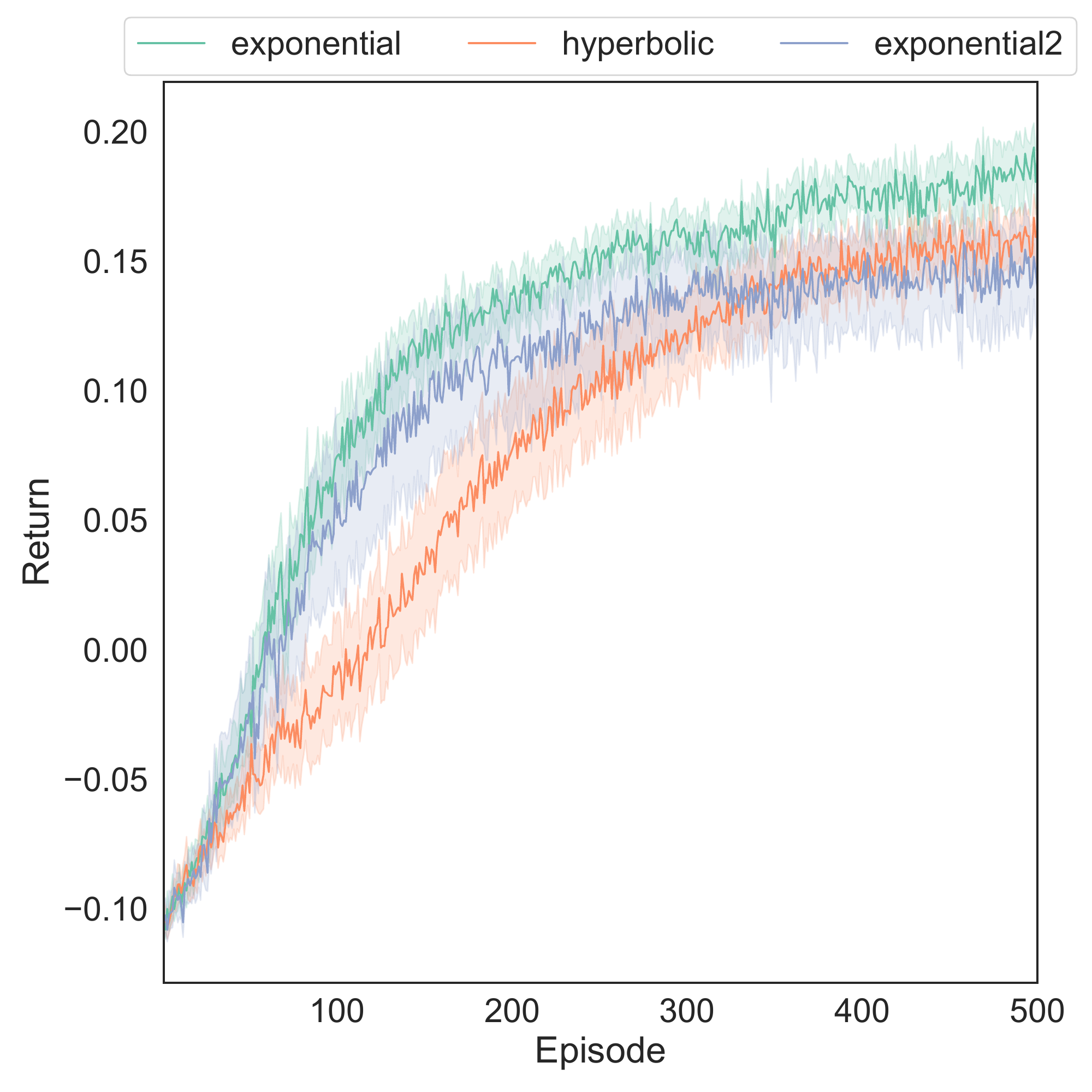}
			{\footnotesize(a) Learning curves of Acrobot}
		\end{minipage}
		\centering
		\begin{minipage}[t]{0.4925\linewidth}
			\centering
			\includegraphics[keepaspectratio=true,width=\linewidth]{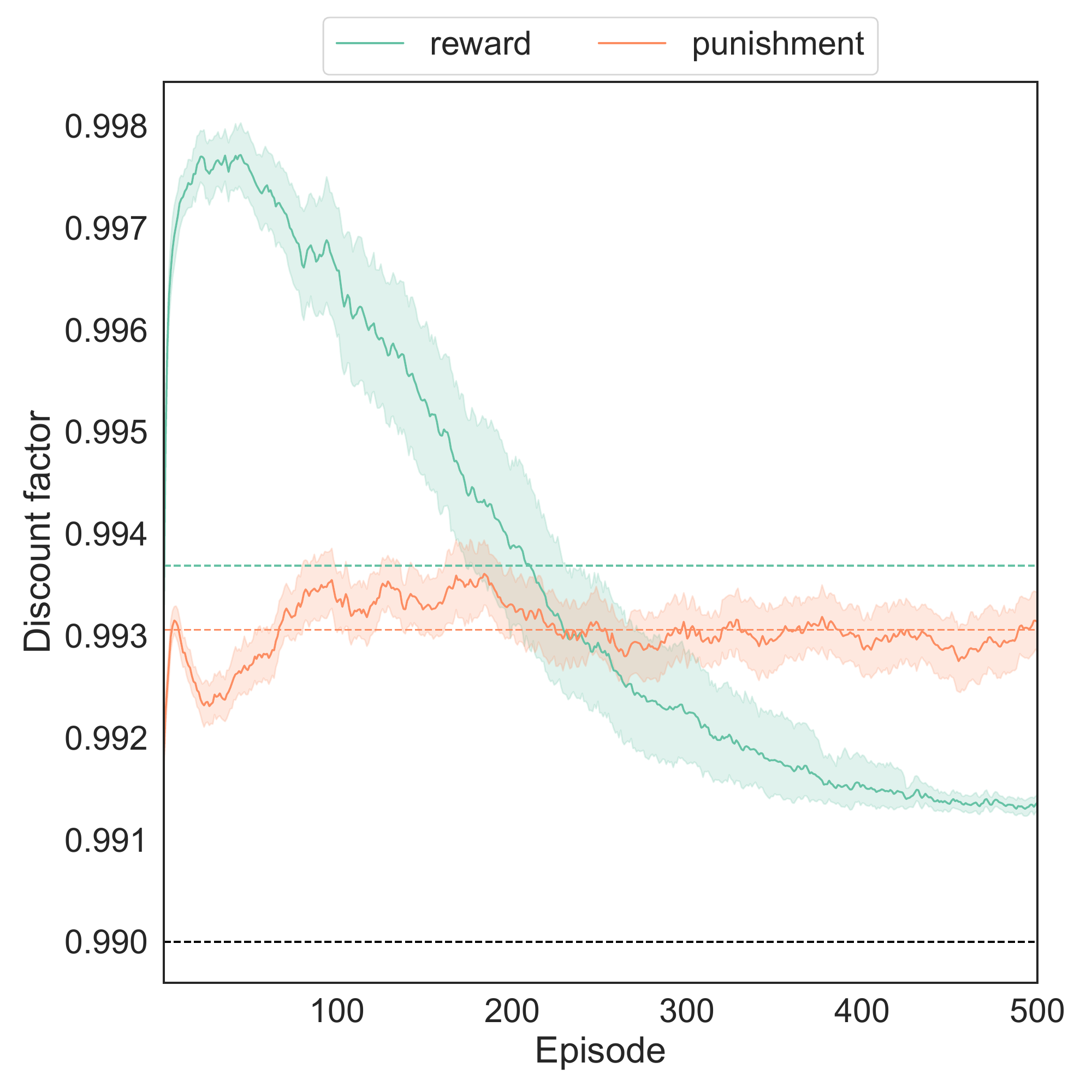}
			{\footnotesize(b) Discount factors of Acrobot}
		\end{minipage}
	\end{minipage}
	\centering
	\begin{minipage}[t]{0.495\linewidth}
		\centering
		\begin{minipage}[t]{0.4925\linewidth}
			\centering
			\includegraphics[keepaspectratio=true,width=\linewidth]{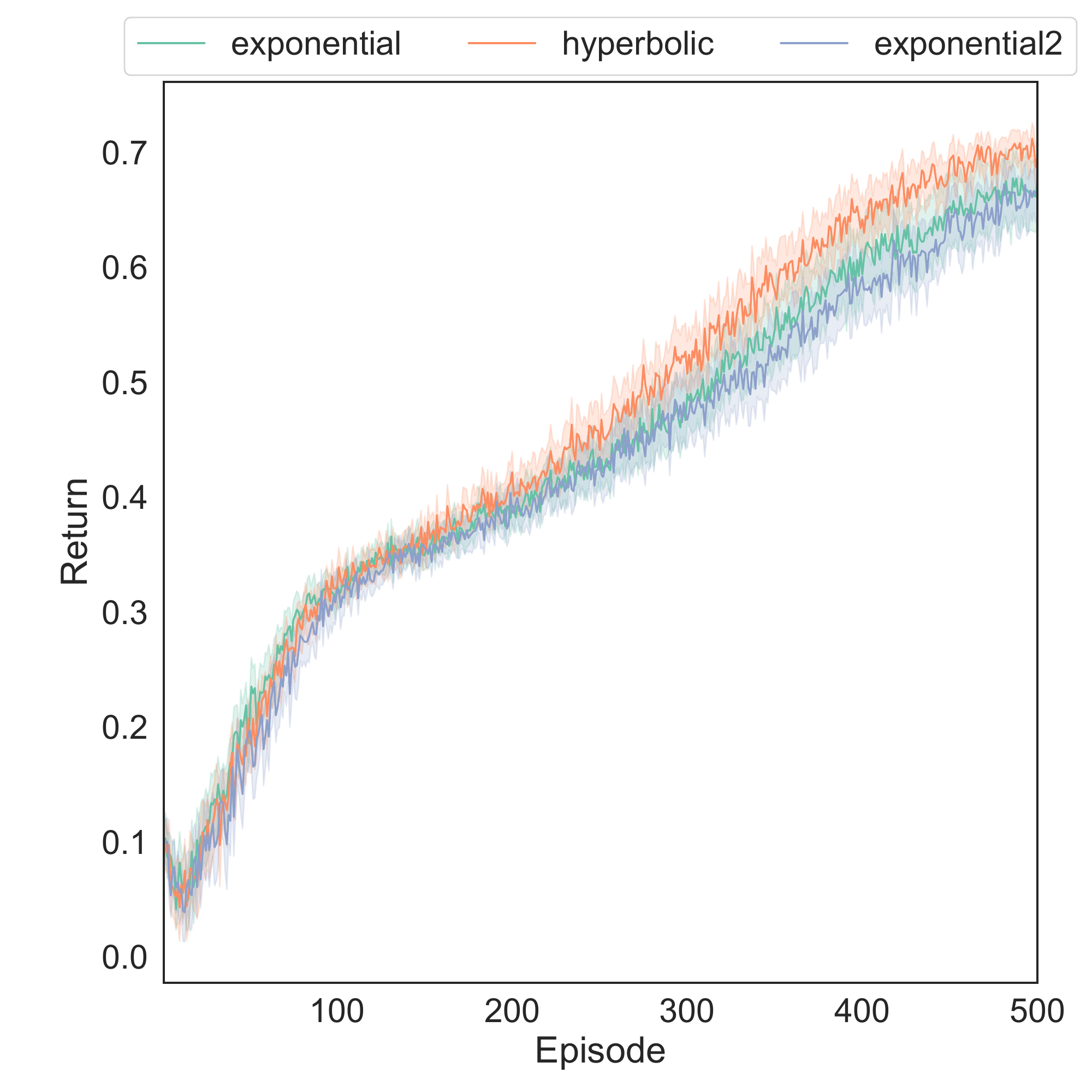}
			{\footnotesize(c) Learning curves of CartPole}
		\end{minipage}
		\centering
		\begin{minipage}[t]{0.4925\linewidth}
			\centering
			\includegraphics[keepaspectratio=true,width=\linewidth]{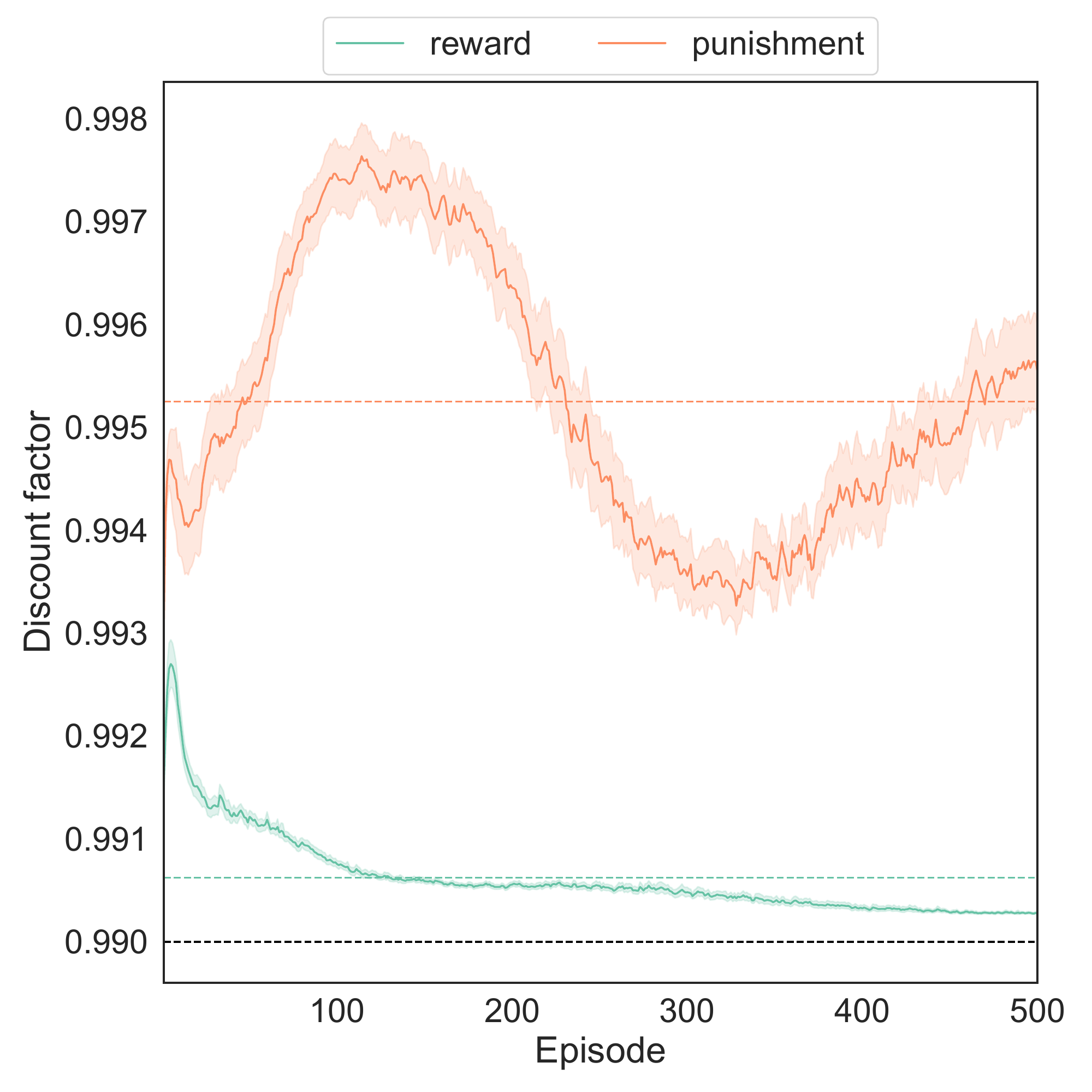}
			{\footnotesize(d) Discount factors of CartPole}
		\end{minipage}
	\end{minipage}
\vspace{-1mm}
\caption{Simulation results}
\label{fig:results}
\end{figure*}

\section{Simulations}

\subsection{Conditions}

In this paper, the performance of RL with the hyperbolic discounting is investigated.
To do so, two environments for numerical simulations are prepared: Acrobot and CartPole (see \url{https://github.com/kbys-t/gym_rp}).
In Acrobot, reward is given only when the target motion is achieved while punishment is continuously given.
On the other hand, CartPole has continuous reward and event-based punishment.
This setting is because, as shown in eq.~\eqref{eq:tde_hyp}, TD error depends on the statics of reward and punishment, which are affected by how to be given.

The hyperparameters $\kappa$, $p$, and $\beta$ are empirically set as $0.01$, $1$, and $0.1$, respectively.
The other parameters for RL is decided to succeeded in learning with the conventional RL.
For comparison, two cases, where the conventional RL has $\gamma_{r,p} = 0.99$ or $\gamma_{r,p}$ the averages of the case with the hyperbolic discounting, are ocnducted.

\subsection{Results}

Fig.~\ref{fig:results} summarized the simulation results of 50 trials per each case.
The case of exonential2 (with the averages of the discount factors in the hyperbolic discounting) had the worst performance in both environments.
Namely, it is expected that the discount factors depending on the value function or state contribute to the learning performance.

The proposal in CartPole outperformed the other cases, although it in Acrobot was inferior to the result of exponential1 (with $\gamma = 0.99$).
Focusing on the average discount factors during each episode (see (b) and (d) in Fig.~\ref{fig:results}), one for reward was likely to be greater than one for punishment in Acrobot; in contrast, their relation was reversed in CartPole.
This difference may be given from the design of reward and punishment: i.e., the continuous design would make the discount factor small; and the event-based design would be a vice versa.
This thought comes from the bias defined in eq.~\eqref{eq:bias_hdtd} as the variance of reward (or punishment), namely, the event-based or sparse design would cause large variance, thereby increasing the discount factor, as expected in eq.~\eqref{eq:tde_hyp}.
Indeed, when learning progressed to some extent, reward was stable gained, so its variance became small, which made its discount factor small accordingly.

If the continuous reward and the event-based punishment are generally better for learning like these results, the asymmetry of the discount factors between them would be expected in this scheme.
That is, if so, the hyperbolically-discounted reinforcement learning on the reward-punishment framework would be related to a sign effect in animals~\cite{tanakaJNS2014}, and would contribute to analyze such animals behaviors mathematically.

\section{Conclusion}

This paper proposed hyperbolically-discounted RL on the reward-punishment framework.
Combining the hyperbolically-discounted TD error and the reward-punishment framework, optimization by RL was enabled.
In simulations, the proposal outperformed the conventional RL, although the performance depends on the design of reward and punishment.
In addition, the averages of discount factors for reward and punishment were in asymmetry, like the sign effect in animal behaviors.

Future work is further analyses of the proposed scheme and to establish the way to design reward and punishment.

%
%
%
%
\bibliographystyle{IEEEtran}
{
\bibliography{hyperbolic}
}

\end{document}